\documentclass{article}
\usepackage[final]{colm2026_conference}

\usepackage[utf8]{inputenc}
\usepackage[T1]{fontenc}
\usepackage{microtype}
\usepackage{booktabs}
\usepackage{multirow}
\usepackage{amsmath,amssymb}
\usepackage{graphicx}
\usepackage{xcolor}
\usepackage{makecell}
\usepackage[colorlinks=true,linkcolor=blue,citecolor=blue,urlcolor=blue]{hyperref}

\newcommand{\gptmini}{GPT-5.4-mini}
\newcommand{\Mr}{M_{\mathrm{r}}}
\newcommand{\Mnr}{M_{\mathrm{nr}}}
\newcommand{\skl}{\sigma}
\newcommand{\qwen}{Qwen3.6-27B}
\newcommand{\ttele}{$\tau^2$-telecom}
\newcommand{\tretail}{$\tau^2$-retail}

\title{Reason Wide, Not Deep: Amortizing the Reasoning Premium into Distilled Skills}

\newcommand{\Author}[1]{\begin{tabular}[t]{c}\textbf{#1}\end{tabular}}

\author{
\makebox[\textwidth][c]{%
  \Author{Agamdeep Singh}\hspace{1.6em}
  \Author{Srishti Gautam}\hspace{1.6em}
  \Author{Priyanshu Gupta}}
\AND
\makebox[\textwidth][c]{%
  \Author{Nikita Mehrotra}\hspace{1.6em}
  \Author{Tanmay Bakshi}\hspace{1.6em}
  \Author{Sumit Gulwani}}
\AND
\normalfont
\makebox[\textwidth][c]{%
  \begin{tabular}[t]{c}
    Microsoft\thanks{Email in order:
      \texttt{\{t-agasingh, srgautam, priyansgupta, nmehrotra, t-tbakshi, sumitg\}@microsoft.com}}
  \end{tabular}}
}

\begin{document}
\maketitle

\begin{abstract}
Reasoning modes of language models outperform their non-reasoning counterparts on multi-step agentic tasks, but pay a 3--6$\times$ premium in output tokens on \emph{every} episode --- much of it spent re-deriving procedures that are shared across episodes of the same domain. We show this recurring cost can be \emph{amortized}: a coding agent analyses a small corpus of existing trajectories from a training split and compiles a compact natural-language \emph{skill} that is injected into the non-reasoning model's system prompt. Across four agentic benchmarks (ALFWorld, $\tau^2$-bench telecom and retail, and SpreadsheetBench-Verified), skills recover 55\%--100\%+ of the reasoning gap for \gptmini{} on held-out tasks --- exceeding the reasoning mode outright on two of four --- while emitting 2.7--6$\times$ fewer output tokens and zero reasoning tokens. Notably, reasoning traces are not a prerequisite: skills distilled from non-reasoning trajectories alone remain competitive with skills distilled from paired reasoning/non-reasoning corpora, with domain-dependent differences between the two sources. We interpret these results through a search lens: test-time reasoning is \emph{deep} search inside a single episode, re-paid at every deployment, while corpus distillation is \emph{wide} search across episodes, paid once. The two recover overlapping procedural knowledge, and width over cheap trajectories is often the better buy --- with the residual gap on some domains (telecom, SpreadsheetBench) delineating where genuinely per-instance deep search remains necessary.
\end{abstract}

\section{Introduction}

\begin{figure}
    \centering
    \includegraphics[width=0.9\linewidth]{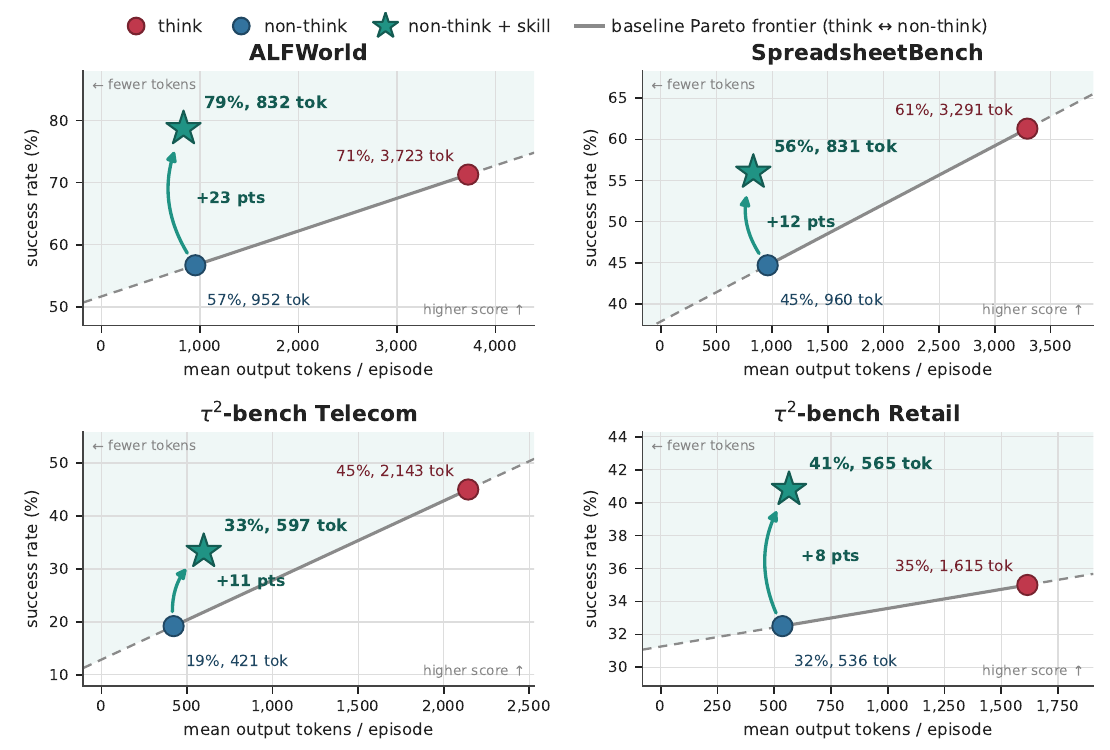}
    \caption{\textbf{Skills break the accuracy--token frontier.} Held-out
    success rate vs.\ mean output tokens per episode for \gptmini{} on four
    agentic benchmarks. The gray line is the baseline
    Pareto frontier traced by toggling the reasoning mode (think $\leftrightarrow$
    no-think); arrows mark the lift from injecting a distilled skill into the
    no-think model. On every benchmark the skill lands above the frontier.}
    \label{fig:mode-comparison}
\end{figure}
Test-time reasoning has become the default recipe for hard tasks: models trained to emit long chains of thought before acting \citep{openai2024o1,guo2025deepseekr1,shao2024deepseekmath} outperform their non-reasoning counterparts on mathematics, coding, and, increasingly, multi-step \emph{agentic} tasks in which the model interleaves tool calls, environment observations, and user turns \citep{yao2023react,shridhar2021alfworld,yao2024taubench,barres2025tau2}. The improvement is real but so is the bill. In our experiments, enabling the reasoning mode of \gptmini{} multiplies per-episode output tokens by 3.0--5.1$\times$ across four agentic benchmarks (up to 6.2$\times$ for \qwen{}), and this premium is paid again on every single episode, forever, because reasoning tokens are generated afresh each time.

Reading the reasoning traces reveals why this is wasteful. Within a fixed domain, much of the deliberation is not instance-specific problem solving but the re-derivation of \emph{episode-invariant procedure}: the retail support agent reasons its way (again) to ``I should not call the account-lookup tool until the customer has actually provided an email''; the household agent re-discovers (again) that ``heat X'' is an atomic command rather than a sequence of microwave-door operations. Non-reasoning rollouts of the same model fail precisely where this procedural knowledge is missing --- in the retail domain, a single recurring bug (calling an authentication tool with a fabricated argument) appears in 59\% of non-reasoning training rollouts and accounts for 94\% of observed tool errors. Recurrent computation is exactly what amortization is for. We ask: \textbf{how much of the reasoning premium can be paid once, offline, instead of on every episode?}

Our method, \emph{passive skill distillation}, is deliberately simple: collect a small corpus of trajectories (35--50 tasks) from a training split, hand the corpus to an off-the-shelf coding agent \citep{claudecode}, and ask it to compile a compact natural-language \emph{skill} --- 40--130 lines of markdown encoding concrete, failure-derived rules --- which is then injected into the system prompt of the \emph{non-reasoning} model. The agent writes and runs its own analysis code over the corpus --- error-type frequencies, action $n$-grams, loop detection, win/loss contrasts --- and compiles what it finds into rules. No weights are updated, no environment rollouts are collected for distillation, no per-instance search is run at deployment, and the skill occupies a cacheable prefix: one pass over logs that production systems already store.

Our contributions are:
\begin{itemize}
\item \textbf{A corpus-to-skill distillation pipeline} requiring only existing rollouts and roughly \$1--\$3 of coding-agent time per domain (Section~\ref{sec:method}).
\item \textbf{Evidence that skills amortize the reasoning premium.} On held-out tasks across ALFWorld, SSB-Verified, and $\tau^2$-bench telecom/retail, skills recover 55\%--100\%+ of the no-think$\to$think gap for \gptmini{}, exceed the reasoning mode outright on ALFWorld and retail, and emit 2.7--5$\times$ fewer output tokens than the reasoning mode with zero reasoning tokens (Section~\ref{sec:results}).
\item \textbf{An ablation on distillation source.} Skills distilled from non-reasoning trajectories alone are competitive with skills distilled from paired think/no-think corpora, with domain-dependent differences in either direction (retail favors the paired corpus, SSB-Verified the no-think-only corpus by 10 points) --- showing that reasoning traces are not a prerequisite for effective distillation (Section~\ref{sec:distill-source}).
\item \textbf{A favorable comparison to automatic prompt optimization.} Against GEPA \citep{agrawal2025gepa}, a state-of-the-art reflective prompt evolver, our distilled skills score higher on both $\tau^2$ domains while costing 4.1$\times$ less to produce (Section~\ref{sec:gepa}).
\end{itemize}

We interpret these results through a search lens (Section~\ref{sec:discussion}): reasoning is \emph{deep} search within one episode; distillation is \emph{wide} search across many. Where the required knowledge is procedural and domain-level, width over cheap trajectories is the better buy. The domains where a residual gap survives (telecom, SSB-Verified) are exactly those where per-instance deliberation --- long dependency chains, instance-specific spreadsheet logic --- cannot be captured by any fixed prompt.

\section{Related Work}

\paragraph{Test-time reasoning and its cost.} Chain-of-thought prompting \citep{wei2022cot} and RL-trained reasoning modes \citep{openai2024o1,guo2025deepseekr1,shao2024deepseekmath} trade tokens for accuracy, and test-time compute can outperform parameter scaling \citep{snell2024scaling,muennighoff2025s1,ye2025limo}. A growing literature documents the inefficiency of this trade --- overthinking on easy instances \citep{chen2024overthinking}, and mitigations via terse drafting or token budgets \citep{xu2025chainofdraft,han2024tokenbudget}. These methods compress reasoning \emph{within} an episode; we amortize it \emph{across} episodes. Whether RL-induced reasoning elicits knowledge already latent in the base model \citep{yue2025limit} is congenial to our finding that the same procedural knowledge can be surfaced by a prompt.

\paragraph{Prompt optimization.} OPRO \citep{yang2024opro}, DSPy/MIPROv2 \citep{khattab2024dspy,opsahlong2024mipro}, TextGrad \citep{yuksekgonul2024textgrad}, and GEPA \citep{agrawal2025gepa} search prompt space against a validation metric, typically via many scored rollouts. Our pipeline is complementary but cheaper in kind: a single reflective pass by a strong coding agent over an \emph{existing} corpus, with no optimization loop. Section~\ref{sec:gepa} compares directly against GEPA.

\paragraph{Experiential learning for agents.} Voyager \citep{wang2023voyager} grows a code skill library online; Reflexion \citep{shinn2023reflexion} feeds verbal self-critique into retries of the \emph{same} task; ExpeL \citep{zhao2024expel} and Agent Workflow Memory \citep{wang2024awm} extract insights or workflows from experience. We share the extract-once-reuse-forever premise, but frame the payoff differently: the skill is a substitute for an expensive \emph{reasoning mode}, evaluated by how much of the think/no-think gap it recovers per token, and produced by an external coding agent rather than by the acting model itself.

\section{Passive Skill Distillation}
\label{sec:method}

\paragraph{Setup.} Let $M$ expose a reasoning mode $\Mr$ (private reasoning tokens before each action) and a non-reasoning mode $\Mnr$ (actions only). A benchmark supplies tasks $\mathcal{T} = \mathcal{T}_{\mathrm{train}} \cup \mathcal{T}_{\mathrm{test}}$ (disjoint), an environment loop (or simulated user), and terminal rewards. The input to distillation is a trajectory corpus $\mathcal{D}$ collected once on $\mathcal{T}_{\mathrm{train}}$: per-step observations, actions and tool calls, visible outputs, and rewards. $\mathcal{D}$ is whatever already exists; no new rollouts are collected for distillation.

\paragraph{Step 1: Collect a training corpus.} For each domain we roll out the model on the training split: 50 ALFWorld canonical tasks, 50 SSB-Verified tasks, 50 \ttele{} and 35 \tretail{} training tasks. In the \emph{paired} condition, $\mathcal{D}$ contains both think and no-think trajectories from the same tasks; in the \emph{no-think-only} condition, only the latter. These are ordinary evaluation rollouts --- in practice such corpora often already exist.

\paragraph{Step 2: Distill with a coding agent.} A coding agent $A$ (an LLM with file-system and code-execution tools; here Claude Code with Claude Sonnet 5 \citep{claudecode}) is opened in the directory containing the corpus and receives a fixed natural-language instruction $P$, producing a skill $\skl = A(\mathcal{D}, P)$. The agent compares failing and succeeding trajectories (and, when available, contrasts no-think failures with think successes on the same tasks), computing corpus-level statistics --- failure-mode frequencies, action loops, win/loss contrasts --- and reading individual episodes where the statistics point. $A$ only reads the trajectory files and mode-level pass rates; it has no environment access. The output is 40--130 lines of markdown whose rules are concrete and traceable to transcript evidence, e.g., from the retail skill: \emph{``Before calling \texttt{find\_user\_id\_by\_email}, check that the customer's message actually contains a real email address \ldots{} this bug appeared in 13 of 22 rollouts and accounted for 17 of 18 tool errors.''} Distillation is a one-time cost of \$1.28--\$2.44 per domain (Section~\ref{sec:gepa}).

\paragraph{Step 3: Deploy.} The skill is appended verbatim to the non-reasoning model's system prompt: $\pi_{\skl}(\cdot) = \Mnr(\cdot \mid \mathrm{sys} \oplus \skl)$. Nothing else --- harness, decoding, tools --- changes between the no-think and skill conditions. The skill adds a fixed, cacheable prompt prefix. Skills are distilled per model and per domain.


\section{Experimental Setup}
\label{sec:setup}

\paragraph{Benchmarks.} \textbf{ALFWorld} \citep{shridhar2021alfworld}: text-based embodied household tasks (ReAct-style agent, admissible commands, max.\ 40 steps); held-out random-50 split; win rate. \textbf{SSB-Verified}: a verified subset of SpreadsheetBench \citep{ma2024spreadsheetbench}, real-world spreadsheet manipulation against live workbooks; held-out 50 tasks; modification accuracy. \textbf{$\tau^2$-bench} telecom and retail \citep{barres2025tau2,yao2024taubench}: conversational customer-service agents with tool use and a simulated user in a dual-control environment; held-out test splits of 40 tasks; pass rate.

\paragraph{Models and modes.} \gptmini{} with \texttt{reasoning\_effort} $\in$ \{none, medium\} and \qwen{} with \texttt{enable\_thinking} $\in$ \{false, true\}, each served through a single gateway so that only the reasoning flag (and, in skill conditions, the system prompt) differs between conditions. Each cell is the mean of 3 evaluation seeds. Skills are produced once per domain per model by Claude Sonnet 5 via Claude Code.

\section{Results and Ablations}
\label{sec:results}

\begin{table}[t]
\centering
\caption{\textbf{Main results.} Held-out success (3 seeds) and mean output tokens per episode for \gptmini{} and \qwen{}. ``Token Reduction'' indicates the token-reduction factor relative to the think mode of the same model and benchmark. Bold marks the best score per benchmark per model; underline marks the second best.}
\label{tab:main}
\small
\begin{tabular}{ll ccc ccc}
\toprule
 & & \multicolumn{3}{c}{\gptmini{}} & \multicolumn{3}{c}{\qwen{}} \\
\cmidrule(lr){3-5}\cmidrule(lr){6-8}
Benchmark & \makecell{\centering Mode} & Score & Tokens & \makecell{Token\\Reduction} & Score & Tokens & \makecell{Token\\Reduction} \\
\midrule
\multirow{3}{*}{ALFWorld}
 & think               & \underline{0.713} & 3{,}723 & --          & 0.773             & 9{,}232 & -- \\
 & no-think            & 0.567             & 952     & 3.9$\times$ & \underline{0.827} & 991     & 9.3$\times$ \\
 & no-think + skill    & \textbf{0.787}    & 832     & 4.5$\times$ & \textbf{0.980}    & 619     & 14.9$\times$ \\
\midrule
\multirow{3}{*}{SSB-Verified}
 & think               & \textbf{0.613}    & 3{,}291 & --          & 0.560             & 2{,}826 & -- \\
 & no-think            & 0.447             & 960     & 3.4$\times$ & \underline{0.640} & 2{,}432 & 1.2$\times$ \\
 & no-think + skill    & \underline{0.560} & 831     & 4.0$\times$ & \textbf{0.673}    & 2{,}729 & 1.0$\times$ \\
\midrule
\multirow{3}{*}{\ttele{}}
 & think               & \textbf{0.450}    & 2{,}143 & --          & \textbf{0.933}    & 6{,}058 & -- \\
 & no-think            & 0.192             & 421     & 5.1$\times$ & \underline{0.883} & 985     & 6.2$\times$ \\
 & no-think + skill    & \underline{0.333} & 597     & 3.6$\times$ & \textbf{0.933}    & 1{,}026 & 5.9$\times$ \\
\midrule
\multirow{3}{*}{\tretail{}}
 & think               & \underline{0.350} & 1{,}615 & --          & \textbf{0.633}    & 4{,}124 & -- \\
 & no-think            & 0.325             & 536     & 3.0$\times$ & \underline{0.600} & 1{,}058 & 3.9$\times$ \\
 & no-think + skill    & \textbf{0.408}    & 565     & 2.9$\times$ & 0.558             & 1{,}180 & 3.5$\times$ \\
\bottomrule
\end{tabular}
\end{table}




\subsection{Skills recover most of the reasoning gap at a fraction of the tokens}
\label{sec:main-results}

\begin{figure}[!t]
    \centering
    \includegraphics[width=1\linewidth]{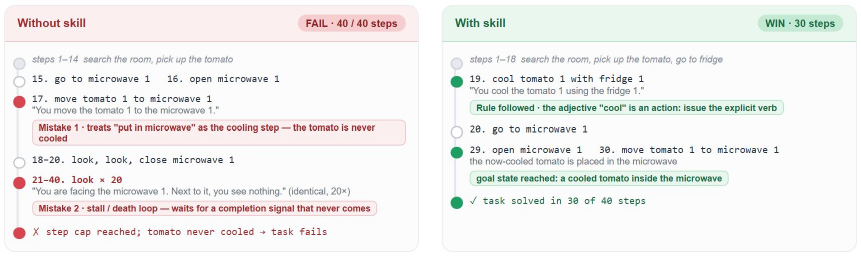}
    \caption{ ALFWorld task -- 
(\emph{``put a cool tomato in microwave''}). Non-reasoning model with and without skill. Without it the model never cools the tomato and loops on \texttt{look} to the step cap
(left); with it the model issues \texttt{cool tomato 1 with fridge 1} and finishes (right). Verbatim from the rollouts.}
\label{fig:alf-traj}
\end{figure}

Table~\ref{tab:main} shows the central result. For \gptmini{}, the reasoning
mode beats no-think on all four benchmarks (by $+14.6$, $+16.6$, $+25.8$, and
$+2.5$ points); injecting a distilled skill into the no-think model recovers
55\%--100\%+ of that gap everywhere, exceeding the reasoning mode outright on
ALFWorld (0.787 vs.\ 0.713) and \tretail{} (0.408 vs.\ 0.350), while emitting
2.9--4.5$\times$ fewer output tokens and zero reasoning tokens. On ALFWorld
and SSB the skill even undercuts the plain no-think baseline in tokens:
fewer flailing retries means shorter episodes (21.8 vs.\ 27.0 turns on
ALFWorld). Because the skill lives in a cacheable system-prompt prefix, its
marginal deployment cost is negligible, while the think premium is re-paid on
every episode. Beating the teacher is not paradoxical: a rule aggregated over
50 training episodes is more reliable than a derivation the reasoning model
must re-produce correctly each time. Indeed, the reasoning model itself
occasionally falls into the ALFWorld appliance-door loop that the skill
forbids outright (Appendix~\ref{app:skills}).

The gains are legible at the level of individual trajectories. Consider a single held-out ALFWorld task, ``put a cool tomato in microwave'' (Figure~\ref{fig:alf-traj}). The non-reasoning baseline picks up the tomato and places it in the microwave without ever cooling it, treating the adjective \emph{cool} as a property rather than a required action; having changed nothing, it then issues \texttt{look} twenty times in a row, waiting for a completion signal that never arrives, and exhausts its 40-step budget. The distilled skill supplies exactly the two missing pieces: a rule that adjectives such as \emph{cool} must be realized with an explicit \texttt{cool X with fridge} command, and a rule to break out of repeated no-op observations. With these, the same model issues the cool command at step 19 and completes the task in 30 steps. These are not isolated fixes: the missed-transform failure occurs in 
35.9\% of transform tasks without the skill and 
11.5\% with it, and stall loops fall from 
28.7\% to 
5.3\%, together accounting for most of the ALFWorld win-rate improvement.

The \qwen{} columns repeat the study with a second model and Qwen-specific
skills. Distillation again helps on three of four benchmarks, reaching 0.980
on ALFWorld (near-ceiling, $+15.3$ over no-think) and 0.673 on SSB-Verified,
and matching the think mode on telecom (0.933) at $5.9\times$ fewer output
tokens, even though \qwen{}'s reasoning mode is itself unreliable and
\emph{hurts} on ALFWorld ($-5.4$) and SSB-Verified ($-8.0$). Retail is the
one regression ($-4.2$ points): with a near-zero think/no-think gap and an
already-competent base, added rules may over-constrain. 

\subsection{Distilling from reasoning v\/s non-reasoning trajectories}
\label{sec:distill-source}

\begin{table}[t]
\centering
\caption{\textbf{Ablation: distillation source.} Held-out pass rate and mean output tokens per episode for a skill distilled from \emph{no-think} rollouts only vs.\ from a paired corpus that additionally includes reasoning (\emph{think}) traces (\gptmini{}); both skills are injected into the same non-reasoning model and emit zero reasoning tokens. Bold marks the higher score per benchmark.}
\label{tab:distill-source}
\small
\begin{tabular}{l cc @{\hskip 1.5em} cc}
\toprule
 & \multicolumn{2}{c}{Think-distilled} & \multicolumn{2}{c}{No-think-distilled} \\
\cmidrule(lr){2-3}\cmidrule(lr){4-5}
Benchmark & Score & Tokens & Score & Tokens \\
\midrule
ALFWorld         & \textbf{0.813} & 748 & 0.787          & 832 \\
SpreadsheetBench & 0.460          & 820 & \textbf{0.560} & 831 \\
\ttele{}         & 0.325          & 533 & \textbf{0.333} & 597 \\
\tretail{}       & \textbf{0.458} & 599 & 0.408          & 565 \\
\bottomrule
\end{tabular}
\end{table}

Our main results (Table~\ref{tab:main}) use skills distilled from non-reasoning
trajectories alone --- the distiller never sees a reasoning trace. A natural
question is whether giving the distiller access to reasoning traces changes the
resulting skill. We therefore ablate the corpus composition for \gptmini{}:
the \emph{no-think-only} condition distills from non-reasoning rollouts, while
the \emph{paired} condition additionally includes think-mode trajectories from
the same training tasks (Section~\ref{sec:method}), allowing the distiller to
contrast no-think failures with think successes on identical tasks.

Table~\ref{tab:distill-source} shows a mixed picture. The two sources are
statistically close on ALFWorld (0.787 vs.\ 0.813) and telecom (0.333 vs.\
0.325). On retail, the paired corpus produces the stronger skill (0.458 vs.\
0.408), suggesting that reasoning traces can supply useful signal --- e.g.,
successful think-mode demonstrations of the authentication discipline that
no-think rollouts consistently violate. On SSB-Verified the ordering reverses,
and sharply: the no-think-only skill scores 10 points higher (0.560 vs.\
0.460). One plausible mechanism is that verbose reasoning narratives anchor
the distiller on what the model \emph{believed} rather than on workbook-level
evidence of what \emph{was true}, but we have not isolated this and note that
each skill was distilled once, so distillation variance is uncontrolled
(Section~7).

We draw two cautious conclusions. First, reasoning traces are not a
\emph{prerequisite} for effective distillation: no-think-only skills are
competitive everywhere and recover 55\%--100\%+ of the reasoning gap in
Table~\ref{tab:main}, which matters practically because the full amortization
loop --- deploy cheap agent $\to$ collect logs $\to$ distill $\to$ redeploy ---
can then run without ever invoking a reasoning model. Second, whether adding
reasoning traces helps or hurts appears to be domain-dependent, and the
per-benchmark differences here are within a range where distillation noise
cannot be ruled out.

\subsection{Comparison with a prompt optimizer}
\label{sec:gepa}

\begin{table}[t]
\centering
\caption{\textbf{Distilled skills vs.\ GEPA-optimized prompts} ($\tau^2$ test splits, \gptmini{} no-think, mean of 3 seeds) Scores and one-time production cost of the two $\tau^2$ skills.}
\label{tab:gepa}
\small
\begin{tabular}{lcccc@{\hskip 1.5em}cc}
\toprule
 & \multicolumn{4}{c}{Pass rate} & \multicolumn{2}{c}{Production cost} \\
\cmidrule(r){2-5}\cmidrule(l){6-7}
Domain & no-skill & think & GEPA & Ours & GEPA & Ours \\
\midrule
\tretail{} & 0.325 & 0.350 & \underline{0.392} & \textbf{0.458} & \$2.26 & \textbf{\$1.28} \\
\ttele{}   & 0.192 & \textbf{0.450} & 0.308 & \underline{0.325} & \$13.02 & \textbf{\$2.44} \\
\bottomrule
\end{tabular}
\end{table}

 We compare against GEPA \citep{agrawal2025gepa}, a
state-of-the-art reflective prompt evolver, on both $\tau^2$ domains
(\gptmini{} rollouts, Claude Sonnet 5 reflection; 120 metric call budget). Table~\ref{tab:gepa}: our distilled skills score
higher on both domains (retail 45.8\% vs.\ 39.2\%; telecom 32.5\% vs.\
30.8\%) at 4.1$\times$ lower production cost (\$3.72 vs.\ \$15.28); the gap
is driven by GEPA's active rollouts for each optimization-proposed prompt. Extending the telecom GEPA budget to 240
metric calls resulted in a byte-identical prompt, indicating  convergence.
\subsection{Robustness}

All results are means over 3 runs with the same protocol for baselines and skills; per-seed numbers show consistent orderings (e.g., Qwen ALFWorld skill: 0.98/0.98/0.98; \gptmini{} telecom skill above no-think on all seeds). Skills were distilled once per domain --- we do not report variance over the distillation itself, a limitation discussed below.

\section{Discussion: Deep vs.\ Wide Search}
\label{sec:discussion}

A unifying reading of Tables~\ref{tab:main}--\ref{tab:gepa} is that test-time reasoning and corpus distillation are two ways of purchasing the same commodity --- procedural knowledge about a domain --- with different cost structures. Reasoning is \emph{deep} search: within a single episode the model explores a tree of considerations before each action. Its knowledge is rediscovered from scratch and its cost recurs per episode. Distillation is \emph{wide} search: many complete trajectories are examined side by side, regularities in the failure distribution are extracted once, and the result is reused for free. When the knowledge that deep search recovers is episode-invariant --- ALFWorld's atomic \texttt{clean/heat/cool} commands, retail's authenticate-then-fetch-orders discipline --- width strictly dominates: it is paid once, and Table~\ref{tab:main} shows it can even exceed think mode, because a rule compiled from 50 episodes is more reliable than a derivation the model must reproduce correctly every time.

The lens also predicts where amortization must fall short. The residual think-over-skill gap on telecom (0.450 vs.\ 0.333) and SSB-Verified (0.613 vs.\ 0.560) marks knowledge that is \emph{not} episode-invariant: telecom tasks hinge on long, instance-specific dependency chains in a dual-control environment (which line, which plan, what the user just toggled), and spreadsheet tasks embed one-off logical structure no fixed prompt anticipates. There, per-instance deep search is doing irreplaceable work, and the two mechanisms are complements: a skill to stop re-buying the invariants, reasoning reserved for the instances that need it.

Finally, the NT-skill result sharpens what distillation actually consumes. Reasoning traces are verbose, stylized, and describe what the model \emph{believed}; environment feedback in failed no-think trajectories records what \emph{was true}. On SSB-Verified the paired-corpus skill (46.0\%) underperformed the no-think-only skill (56.0\%), consistent with the distiller anchoring on reasoning narratives instead of workbook-level failure evidence. Wide search needs breadth of outcomes, not depth of introspection --- which is convenient, since non-reasoning trajectories are the cheap ones.

\paragraph{Amortization economics.} Distillation is one coding-agent pass over the corpus (\$1.28--\$2.44 per domain; Table~\ref{tab:gepa}). Per episode, the skill then saves $\Delta = T_{\mathrm{think}} - T_{\mathrm{skill}}$ output tokens --- e.g., $2{,}143 - 597 = 1{,}546$ on telecom, essentially the model's entire 1,572-token reasoning budget --- while adding only a fixed, cacheable input prefix. The one-time cost is repaid once cumulative per-episode savings exceed it; every subsequent episode is pure savings. By contrast, active prompt optimizers \citep{agrawal2025gepa,yang2024opro} spend an evaluation-rollout budget \emph{before} any savings accrue, and cannot run at all where fresh rollouts are unavailable.

\paragraph{Relation to elicitation.} RL-trained reasoning appears to \emph{elicit} latent base-model capabilities rather than create new ones \citep{yue2025limit,muennighoff2025s1,ye2025limo}. Our results are the prompt-side counterpart: if the non-reasoning model already carries the priors needed to execute winning procedures, a distilled description of where search reliably lands is a sufficient --- and far cheaper --- elicitor. The corpus-source ablation sharpens this: even the description need not come from the reasoning model.

\section{Limitations}

Skills were distilled once per model--domain pair; we measure evaluation variance (3 seeds) but not distillation variance, and the Qwen retail regression suggests the process is not uniformly reliable. Results cover two models and four domains; skills are model-specific and cross-model transfer is untested.

\section{Conclusion}

A small corpus of ordinary trajectories, one pass by a coding agent, and a hundred lines of markdown recover most --- sometimes all --- of what an expensive reasoning mode buys on agentic benchmarks, at 2.7--6$\times$ fewer output tokens per episode and a one-time cost of a few dollars --- and the corpus need not contain a single reasoning trace. Reasoning re-derives domain procedure inside every episode; distillation extracts it once. 

\bibliographystyle{plainnat}
\bibliography{references}

\clearpage
\appendix

\section{Distilled Skill Excerpts}
\label{app:skills}

Abridged excerpts from the distilled skills (full files range from 38 to 126 lines of markdown). Rules are imperative, concrete, and cite corpus statistics computed by the distiller.

\paragraph{ALFWorld (no-think-distilled), Rule 1 of 5.}
\begin{quote}\small
\emph{Adjectives in the task are actions, not descriptions.} If the task says ``clean'', ``hot'', or ``cool/cold'' X, that adjective is a required state-change step. You must issue the explicit verb: \texttt{clean <object> with sinkbasin 1}, \texttt{heat <object> with microwave 1}, \texttt{cool <object> with fridge 1}. Do not substitute ``open microwave, move object in, close, open, take back out'' for \texttt{heat X with microwave 1} --- opening/closing an appliance does not perform the transformation.
\end{quote}

\paragraph{\tretail{} (paired-corpus), Rule 1 of 6.}
\begin{quote}\small
\emph{Never call an authentication tool with a guessed or placeholder argument.} This was the single most common bug: it appeared in 13 of 22 rollouts (59\%) and accounted for 17 of 18 tool errors observed (94\%). Before calling \texttt{find\_user\_id\_by\_email} or \texttt{find\_user\_id\_by\_name\_zip}, check that the customer's message actually contains a real email address, or a real first name $+$ last name $+$ zip. If none is present yet, do not call any lookup tool --- respond in plain text asking for one.
\end{quote}

\paragraph{SSB-Verified (no-think-distilled), central rule.}
\begin{quote}\small
\emph{Finish inside the workbook, not just in the chat.} Never end a task by only describing a formula, a macro, or an approach in your response --- explaining the right formula and then not entering it into the sheet is the single most common way this task goes wrong. If the user asks for a VBA macro, still apply the equivalent transformation directly to the workbook via code.
\end{quote}


\section{Reproduction Details}
\label{app:repro}

\paragraph{Splits.} ALFWorld: distill on 50 training tasks, evaluate on a
disjoint held-out set of 50 tasks. $\tau^2$: distill on 50
(telecom) / 35 (retail) training tasks, evaluate on the provided 40-task test
splits. SpreadsheetBench-Verified: distill on 50 training tasks,
evaluate on a disjoint held-out set of 50 tasks. For ALFWorld and
SSB-Verified, which do not ship a designated test split, the held-out sets are
sampled once and fixed across all conditions.

\paragraph{Harness.} ALFWorld uses a ReAct-style agent with
admissible-command grounding and a 40-step cap. SSB-Verified uses a
ReAct-style tool-use harness in which the model writes and executes
\texttt{openpyxl} and \texttt{pandas} code against a live copy of the
workbook, following the original SpreadsheetBench setup
\citep{ma2024spreadsheetbench}. $\tau^2$ uses the standard runner with an LLM
user simulator. In all cases the skill is injected by appending it to the
agent's system prompt with no other change to harness, decoding, or tools;
the same protocol is used for all conditions.


\end{document}